\documentclass[runningheads]{llncs}
\usepackage{graphicx}
\usepackage{comment}
\usepackage{amsmath,amssymb}
\usepackage{color}
\usepackage{makecell}
\begin{document}
\title{hSDB-instrument: Instrument Localization Database for Laparoscopic and Robotic Surgeries}
\titlerunning{hSDB-instrument}
%
\author{Jihun Yoon\inst{1 \bigoplus} \and
Jiwon Lee\inst{1} \and
Sunghwan Heo\inst{1} \and
Hayeong Yu\inst{1} \and
Jayeon Lim\inst{1} \and \\
Chi Hyun Song\inst{1} \and
SeulGi Hong\inst{1} \and
Seungbum Hong\inst{1} \and
Bokyung Park\inst{1} \and \\
SungHyun Park\inst{2} \and
Woo Jin Hyung\inst{1,2} \and
Min-Kook Choi\inst{1 \bigotimes}
}

\authorrunning{J. Yoon et al.}
\institute{hutom, Seoul, Republic of Korea\\
\email{yjh2020@hutom.io\inst{\bigoplus}},
\email{mkchoi@hutom.io\inst{\bigotimes}} \and
Department of Surgery, Yonsei University College of Medicine,\\Seoul, Republic of Korea}
%
%

\maketitle              
\begin{abstract}
Automated surgical instrument localization is an important technology to understand the surgical process and in order to analyze them to provide meaningful guidance during surgery or surgical index after surgery to the surgeon. We introduce a new dataset that reflects the kinematic characteristics of surgical instruments for automated surgical instrument localization of surgical videos. The hSDB(hutom Surgery DataBase)-instrument dataset consists of instrument localization information from 24 cases of laparoscopic cholecystecomy and 24 cases of robotic gastrectomy. Localization information for all instruments is provided in the form of a bounding box for object detection. To handle class imbalance problem between instruments, synthesized instruments modeled in Unity for 3D models are included as training data. Besides, for 3D instrument data, a polygon annotation is provided to enable instance segmentation of the tool. To reflect the kinematic characteristics of all instruments, they are annotated with head and body parts for laparoscopic instruments, and with head, wrist, and body parts for robotic instruments separately. Annotation data of assistive tools (specimen bag, needle, etc.) that are frequently used for surgery are also included. Moreover, we provide statistical information on the hSDB-instrument dataset and the baseline localization performances of the object detection networks trained by the MMDetection library and resulting analyses\footnote{The dataset, additional dataset statistics and several trained models are publicly available at \url{https://hsdb-instrument.github.io/}}.

\keywords{Surgical Instrument Localization, Object Detection, Class Imbalance, Domain Randomization}
\end{abstract}

\section{Introduction}
In the last 30 years, the number of open surgeries has decreased significantly, and the number of minimally invasive surgeries that can speed up recovery by minimizing patient complications has increased significantly. In particular, minimally invasive surgeries using robots are increasing more rapidly in terms of the convenience of the surgical procedure and the minimization of deviations of surgical performances among specialists \cite{Hughes-Hallett15,Perez19}. Owing to the nature of laparoscopic surgery, which is performed by a specialist while looking at the inside of the patient, minimally invasive surgery is recorded in the form of a video that covers all the operations taking place inside the patient. If the operation process in the vidoe is recognized, the operation can be automatically evaluated and analyzed \cite{Jin18}. Furthermore, it may be possible to automate certain surgical procedures, like autonomous driving, as well as real-time guidance during operations, like ADAS (Advanced Driver Assisted System) in vehicle driving.

\begin{table}[t!]
  \caption{\textbf{Summary of datasets for surgical instruments recognition} Surgery type, recognition type, the number of surgeries, the number of instruments, the number of frames and if the synthetic data was included were investigated. The additional frames of synthetic and domain randomization data are highlighted in bold.}
  \label{tab1}
  \resizebox{\linewidth}{!}{
  \begin{tabular}{c|c|c|c|c|c|c} \hline
     & \makecell{Surgery\\Type} & \makecell{Recognition\\Type} & \makecell{\# of\\Surgeries}  & \makecell{\# of\\Instruments} & \makecell{\# of\\Frames} & \makecell{Synthetic\\Data}\\ \hline
 	\cite{Maier-Hein21}	& \makecell{Laparoscopic\\colorectal surgeries} & \makecell{Instance\\segmentation} & 30 cases  & 7 laparoscopic& 10,040 & X\\ \hline
    \cite{Twinanda17} & \makecell{Laparoscopic\\cholecystectomy} & \makecell{Tool\\presence} & 80 cases  & 7 laparoscopic& 176,020 & X\\  \hline 
 	\cite{Allan19} & \makecell{Robotic abdominal\\porcine procedures} & \makecell{Binary(parts)\\segmentation} & 10 cases  & 7 robotic& 600 & X\\ \hline
 	\cite{Sarikaya17} & \makecell{Robotic\\surgery(skill training)} & \makecell{Object\\detection}& \makecell{99 cases for\\6 surgical tasks} & 1 robotic& 22,467 & X\\ \hline
 	\makecell{hSDB-\\instrument} & \makecell{Laparoscopic\\cholecystectomy\\and robotic\\gasterctomy}& \makecell{Object\\detection}& \makecell{ 24 cases for\\cholecystectomy\\and 24 cases for\\ gastrectomy} & \makecell{10 laparoscopic,\\6 robotic and\\4 assistive} &\makecell{26,919(\textbf{+12,428})\\for cholecystectomy\\and 42,891(\textbf{+13,247})\\for gastrectomy} & O\\ \hline
\end{tabular}}
\end{table}

To analyze and evaluate the surgical procedures by using surgical videos in an automated manner, localization, and motion estimation of the surgical tools are essential. To this end, several research teams have released datasets for instrument localization of laparoscopic instruments for colorectral surgeries \cite{Maier-Hein21}, cholecystectomy \cite{Twinanda17} and of robotic  instruments for abdominal porcine procedures \cite{Allan19} and robotic surgery skill training \cite{Sarikaya17}. However these datasets did not provide sufficient annotations for pixel-level instrument localization and various surgical instruments and did not handle the class imbalance problem. In the computer vision community, with the help of publicly available datasets such as MS-COCO \cite{Lin14} and KITTI \cite{Geiger12}, visual recognition techniques for object detection have been developed to a very high level \cite{Ren15,Liu18,Lin17-1}. To achieve a similar improvement in surgical instrument localization, large-scale datasets that contain a wider range of localization information about surgical tools are necessary.

To address this need, we present a hSDB(hutom Surgery DataBase)-instrument dataset for instrument localization in laparoscopic and robotic surgery. The hSDB-instrument dataset is produced from real surgical situation videos, and it contains the localization information of tools used in the surgical procedure. The hSDB-instrument contains instrument information for a total of 24 cases of laparoscopic cholecystectomy and 24 cases of robotic gastrectomy for gastric cancer. Localization information of the surgical instruments is provided in the form of a bounding box to train object detection networks, and each instrument's bounding box is divided into head, wrist, and body parts according to the kinematic characteristics of each tool. The information also includes needles, specimen bags, and surgical tubes that are often used in the surgical process, even if they are not laparoscopic or robotic surgical instruments. Besides, the robotic surgery includes several assistive laparoscopic instruments that are frequently used for gastrectomy.

Moreover, the hSDB-instrument dataset provides localization information on virtual surgical instruments made by Unity and real tools from real minimally invasive surgery videos. There is a severe class imbalance between the surgical tools in the real surgeries. To resolve the class imbalance, we provide localization annotations in two-dimensional (2D) projection images of three-dimensional (3D) models of several laparoscopic and robotic surgical instruments. In the case of synthetic tools, the annotations are free and  include both pixel-level information and bounding box. The synthetic data can be used for training the recognition models by itself, and can also be used for training by applying techniques such as image-to-image translation using generative adversarial neural networks that were recently proposed \cite{Huang18,Lee19,Park19,Pfeiffer19}. We also provide baseline performances for object detection models for the hSDB-instrument dataset, which is expected to help develop new models and algorithms for instrument localization. The models were trained by the MMDetection library \cite{Chen19} based on PyTorch \cite{Paszke19}. \\

The technical contributions of the hSDB-instrument dataset are:
\begin{itemize}
  \item The hSDB-instrument dataset was created based on the instrument's localization information of real surgical procedures. The datasets were split into training and validation sets by each case of surgeries, not just the number of annotations. This means that the generalization error of the recognition models according to different patients' condition is considered.\\
  \item We provide synthetic datasets to handle class imbalance that may occur during data acquisition. Two-dimensional projected pixel-based annotations of 3D synthetic models for several instruments used in the surgical procedure are provided. Several approaches can be employed to address class imbalance using the annotations of the synthetic data. \\
  \item The hSDB-instrument dataset provides baseline performances in instrument localization using MMDetection-based model evaluations. We expect that the baseline models can be used as a guideline in developing new surgical tool localization models or algorithms.
\end{itemize}
\begin{figure*}[t!]
\begin{center}
\includegraphics[width=0.9\textwidth]{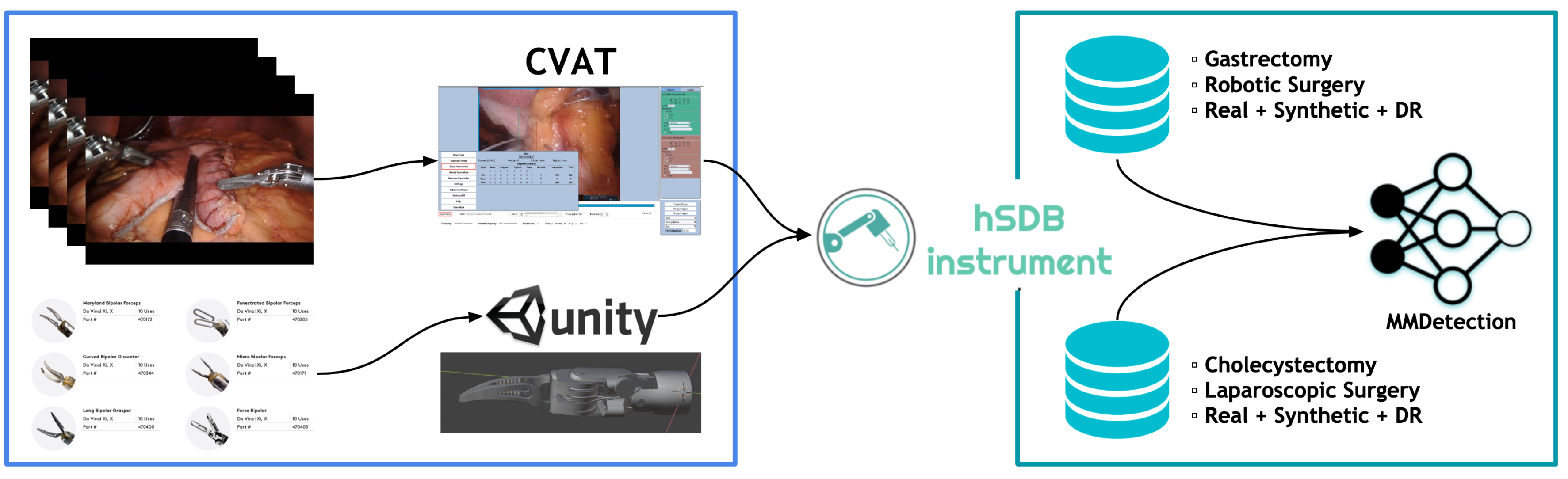}
\caption{\textbf{Schematic representation of data acquisition and storage of the hSDB-instrument dataset.} The box on the left shows the data acquisition process, and the box on the right shows the type of stored data. The hSDB-instrument dataset provides a comparative evaluation of the baseline models trained by using the MMDetection library from the training set.} \label{fig1}
\end{center}
\end{figure*}

\begin{figure*}[t!]
\begin{center}
\includegraphics[width=0.9\textwidth]{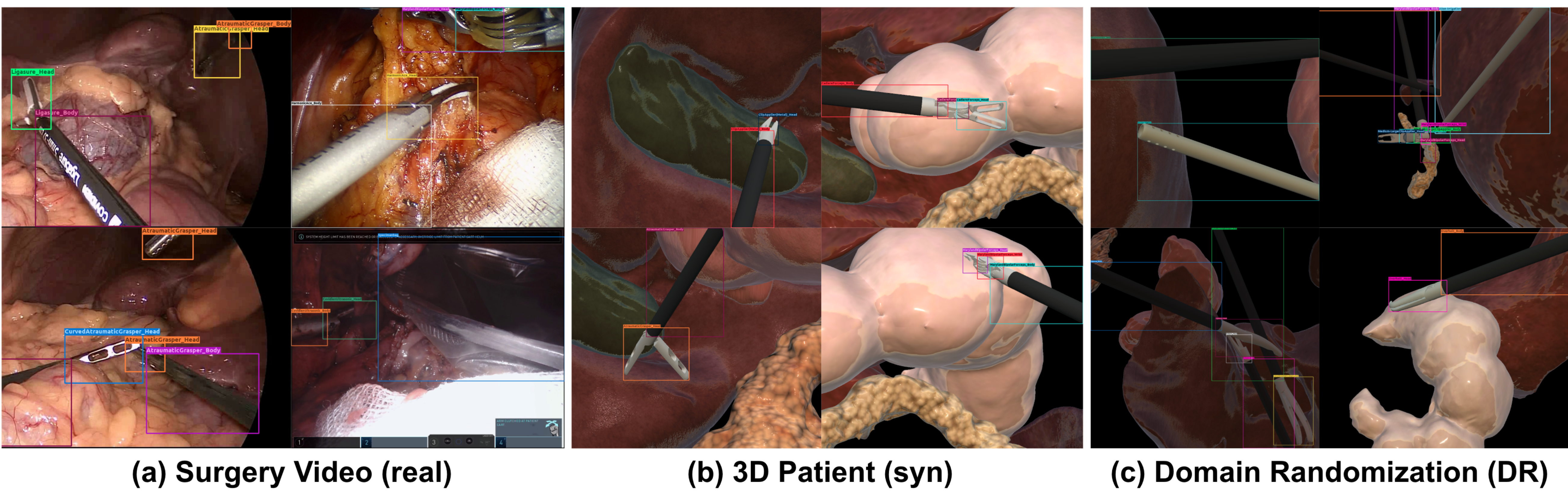}
\caption{\textbf{Annotation visualization using hSDB-instrument API.} (a) shows bounding boxes of tools with respect to a frame extracted from a video. (b) shows the localization information of the tools using the synthetic data from patient CT (Computer Tomography), and (c) shows the localization information of the data generated by the domain randomization method.} \label{fig2}
\end{center}
\end{figure*}

\section{Data Collection}
Figure \ref{fig1} shows the data acquisition platform and storage structure of the hSDB-instrument dataset. In both cholecystectomy and gastrectomy, the format of the annotation database follows the basic structure of the MS-COCO dataset \cite{Lin14}. This section describes the details of the annotation generation tool and method used in the data acquisition process shown on the left side of Figure \ref{fig1}.

\subsection{Laparoscopic cholecystectomy}
The original videos of cholecystectomy included in the hSDB-instrument dataset were a subset of the Cholec80 dataset \cite{Twinanda17}. A web-based computer vision annotation tool called CVAT \cite{cvat} was used to annotate surgical instruments in the form of a bounding box. All the tools were divided into the head and body structures according to the kinematic characteristics of laparoscopic tools. Annotations for all tools were sampled at an average of 1 frame per second. All the annotations were marked by two trained annotators, cross-validated with each other, and only the annotations finally approved by a supervisor, medical expert were stored in the database.

The synthetic dataset was generated by two types of 3D modeling. The first is a synthetic dataset with a 3D environment assuming the real surgical procedure, near the gallbladder modeled with the CT (Computer Tomography) data of a specific patient and the laparoscopic tool. Similar to the robotic surgery console interface, the user inputs commands for the operation of the tool to generate the data while the user interacts with the synthetic environment inside the patient. The second is a synthetic dataset generated by a method so-called domain randomization \cite{Tremblay18}. Domain randomization-based data were randomly generated within a set of constraints of all the tools and organs in the 3D synthetic environment, including location, camera type, lighting, viewpoint, and color of objects. All the 3D modeling data was produced by Unity, and the 3D model information was projected as 2D information corresponding to a specific camera viewpoint and stored as the final training data.

\subsection{Robotic gastrectomy for gastric cancer}
The original videos of gastrectomy for gastric cancer, included in the hSDB-instrument dataset, consist of 24 robotic surgery recordings by experienced specialists. The da Vinci Si system was used to record 14 videos and the da Vinci Xi system was used for the rest 10 of videos. Gastrectomy was performed by an experienced medical specialist, and each time of the recorded videos was between 1.5 and 4 h, depending on types of surgery. As in the case of cholecystectomy, 1 f/s sampling was applied to all the frames in which the tool appeared to obtain annotated information.

Unlike the laparoscopic instruments, the robotic instruments in gastrectomy were annotated in three parts according to their kinematic characteristics. In addition to robotic surgical instruments, it includes several hand-assisted laparoscopic instruments and assistive tools. In the case of the 3D modeling data, the CT data of the cholecystectomy patient were used, and the two types of 3D training data (surgical environment and domain randomization) for robotic surgery instruments were generated in the same manner as that for the laparoscopic surgery. Figure \ref{fig2} shows the visualization of our dataset.


\section{Statistics of hSDB-instrument dataset}
\begin{figure*}[t!]
\begin{center}
\includegraphics[width=1.0\textwidth]{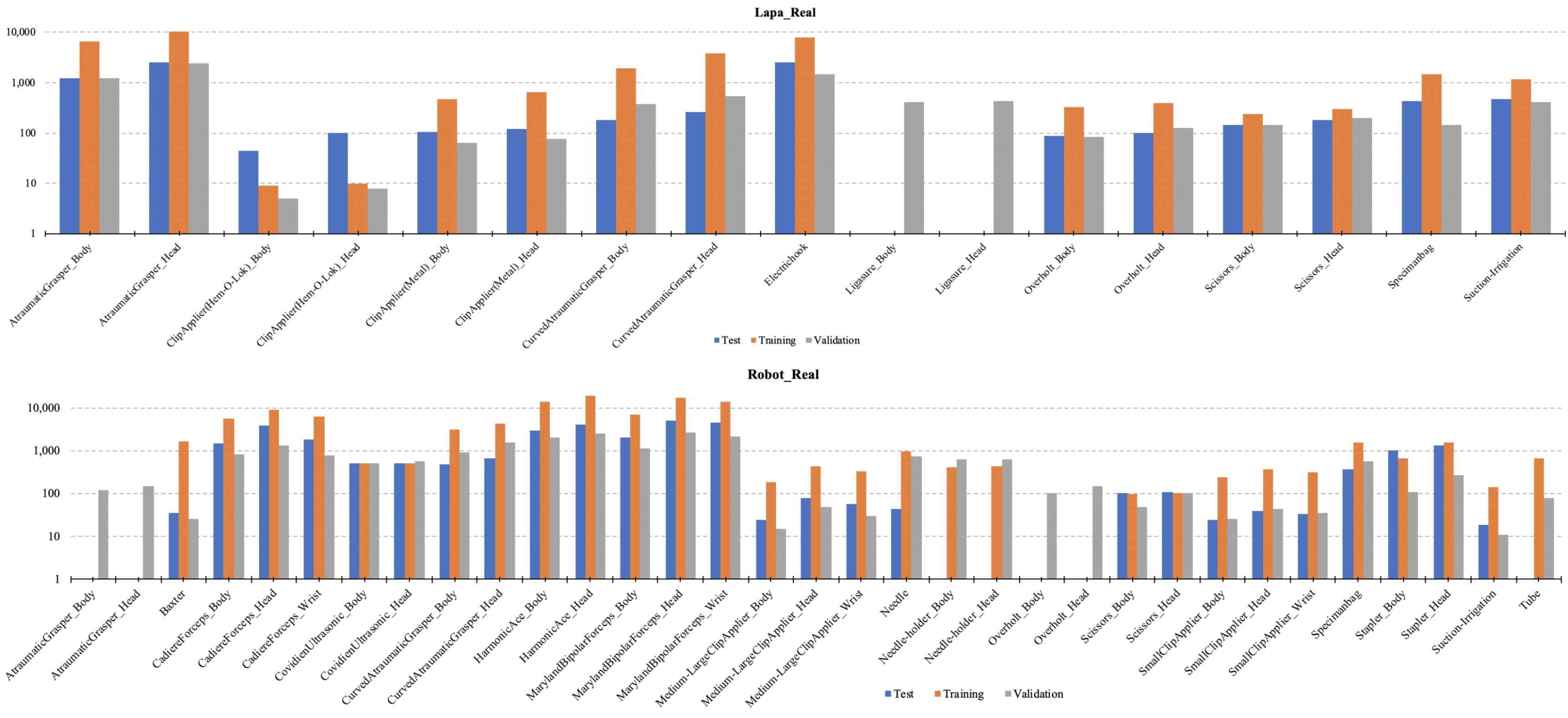}
\caption{\textbf{Distribution of annotations obtained from laparoscopic cholecystectomy and gastrectomy for gastric cancer videos.} The number of annotations is adjusted on a logarithmic scale. Both cholecystectomy and gastrectomy have severe class imbalance problems with several instruments.} \label{fig3}
\end{center}
\end{figure*}

The annotations for each of the 24 surgery cases for both cholecystectomy and gastrectomy have a different distribution of instrument quantities. Figure \ref{fig3} shows a distribution of annotations for the surgical instruments in the hSDB-instrument dataset. 18 of the 24 surgery cases were used to generate training data, 3 cases for validation data, and the last 3 cases for test data assuming unseen surgical situations. 10 types of laparoscopic tools are used for cholecystectomy—atraumatic grasper, clip applier (ham-o-lock), clip applier (metal), curved atraumatic grasper, electrichook, ligasure, overholt, scissors, specimen bag, and suction-irrigation. In the case of gastrectomy, a total of 17 surgical instruments are used, including robotic tools, laparoscopic tools, and auxiliary tools. atraumatic grasper, baxter, cadiere forceps, covidien ultrasonic, curved atraumatic grasper, harmonic ace, maryland bipolar forceps, medium-large clip applier, needle, needle holder, overholt, scissors, small clip applier, stapler, specimen bag, suction-irrigation, and tube.

\begin{figure*}[t!]
\begin{center}
\includegraphics[width=1.0\textwidth]{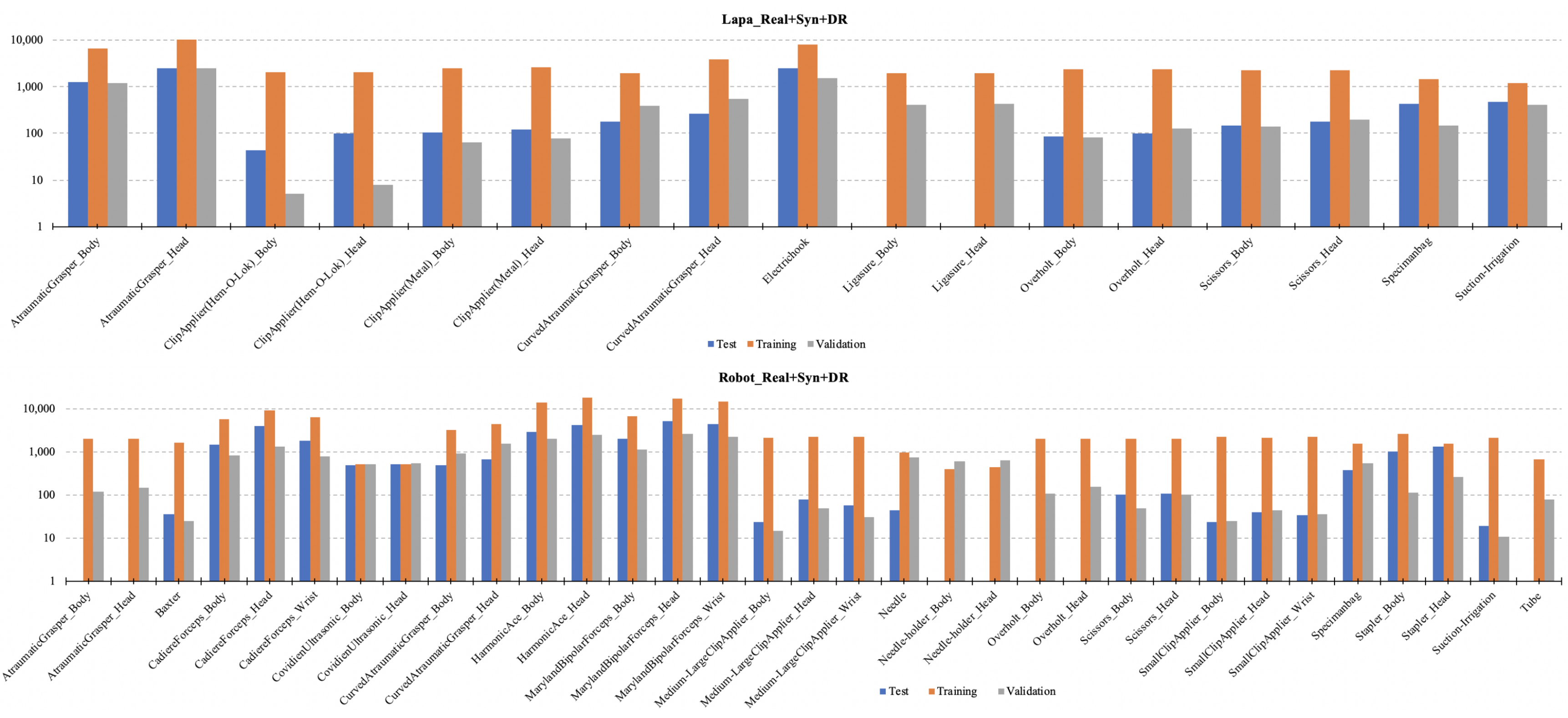}
\caption{\textbf{Distribution of annotations obtained from laparoscopic cholecystectomy and gastrectomy for gastric cancer videos and 3D patient model.} Compared to the distribution of annotations obtained from only the surgery videos (real), the class imbalance problem is alleviated in both surgeries in the real+synthetic+domain randomization data distribution.} \label{fig4}
\end{center}
\end{figure*}

In the case of the synthetic data, it can be randomly generated regardless of data distribution. Therefore, we generated a uniform amount of synthetic data for some of the tools available for 3D models. Next, the available 3D models include 11 tools for gastrectomy (atraumatic grasper, cadiere forceps, curved atraumatic grasper, harmonic ace, maryland bipolar forceps, medium-large clip applier, overholt, scissors, small clip applier, stapler, and suction-irrigation) and 9 laparoscopic tools (atraumatic grasper, clip applier (hem-o-lok), clip applier (metal), curved atraumatic grasper, electrichook, ligasure, overholt, scissors, and suction-irrigation). Two different types of training data were generated with different distributions of synthetic data. First type is training data including the real data and synthetic data with the same amount of each instrument, regardless of a class distribution of the real data to identify the effect of the synthetic data on training. Second type is training data in which synthetic data is added to the real data in order to make up for the lack of annotations for each instrument and verify if the virtual data can solve class imbalance problem. Figure \ref{fig4} shows the distribution of the real and synthetic training data for the two types of data\footnote{Additional statistics is described at https://hsdb-instrument.github.io}.

\section{Baseline Localization Performances}

We trained CNN-based object detection models with MMDetection library \cite{Chen19} to provide the baseline performances of the hSDB-instrument dataset. We evaluated the mAP averaged for IoU $\in$ [0:5 : 0:05 : 0:95] (COCO’s standard metric) where APs were computed at 10 IoU thresholds. The AP is the average of precision $P(\beta)$ over the different levels of recall $R(\beta)$ achieved by varying a confidence threshold $\beta$. The mAP is the mean of APs over all object categories. We trained representative two-stage models: Faster R-CNN \cite{Ren15}, Cascade R-CNN \cite{Cai18} and one-stage models: RetinaNet  \cite{Lin17-2}, FCOS \cite{Tian19}, FoveaBox \cite{Kong19}. The trained models also included feature pyramid structure \cite{Lin17-1}, Libra module \cite{Pang19}, double head module \cite{Li19}, and several  backbones \cite{He16,Xie17,Wang19} with each model. Table \ref{tab2} summarizes the performance changes depending on each detector and its associated submodules\footnote{Additional experiment details and results are described at \\ https://hsdb-instrument.github.io/}. And table \ref{tab3} shows how our synthetic data helps models to be improved in both types of surgeries.

\begin{table}[t!]
  \caption{\textbf{Performance evaluation result for validation set using hSDB-instrument dataset.} Each object detection model combines with a submodule and backbone to show different performances. We evaluated the mAP averaged for IoU $\in$ [0:5 : 0:05 : 0:95]}
  \label{tab2}
  \resizebox{\linewidth}{!}{
  \begin{tabular}{c|c|c|c} \hline
    \multicolumn{4}{c}{Laparoscopic cholecystectomy (real)}\\ \hline 
Model 		& Backbone& Submodule & mAP $[0.5:0.05:0.95]$\\ \hline 
Faster R-CNN & ResNet50 & FPN & 20.2 \\ 
			& 		 & FPN+OHEM & 20.1 \\ 
			& 		 & FPN+Double head & 22.9 \\
 			& ResNet101 & FPN & 22.8  \\  
 			& ResNeXt101-32x4d & FPN & 24.1 \\ 
 			& ResNeXt101-64x4d & FPN & 24.9 \\  
			& HRNetV2p-W18 & - & 23.4 \\ 
			& HRNetV2p-W40 & - & 24.1 \\ \hline 
Cascade R-CNN & ResNet101 & FPN & 24.7  \\
			  & ResNeXt101-64x4d & FPN & 24.5 \\
			& HRNetV2p-W32 & - & \textbf{25.7} \\ \hline
FCOS 		& HRNetV2p-W32-GN & FPN & 11.8  \\
			  & ResNeXt101-64x4d & FPN+mstrain & 22.0  \\ \hline 
FoveaBox & ResNet50 & FPN & 22.9 \\
 		& ResNeXt101 & FPN+align-gn-ms & 23.7  \\  \hline \hline 		
    \multicolumn{4}{c}{Robotic gastrectomy for gastric cancer (real)}  \\ \hline 
Model 		& Backbone& Submodule & mAP $[0.5:0.05:0.95]$\\ \hline 
Faster R-CNN & ResNet50 & FPN+Double head & 37.9 \\ 
			& HRNetV2p-W40 & - & 37.8 \\ \hline
Cascade R-CNN & ResNet101 & FPN & 38.2  \\
			  & ResNeXt101-64x4d & FPN & \textbf{39.9}  \\ 
			& HRNetV2p-W32 & - & 38.8 \\ \hline
FoveaBox & ResNeXt101 & FPN+align-gn-ms & 37.5  \\  \hline
\end{tabular}}
\end{table}

\begin{table}[t!]
  \caption{\textbf{Performance evaluation result for validation set using the second type of training set.}  We evaluated the mAP averaged for IoU $\in$ [0:5 : 0:05 : 0:95].}
  \label{tab3}
  \resizebox{\linewidth}{!}{
  \begin{tabular}{c|c|c|c|c|c|c}
\hline
\multicolumn{3}{c|}{Laparoscopic cholecystectomy} & \multicolumn{4}{c}{mAP $[0.5: 0.05: 0.95]$} \\ \hline
Model 		& Backbone& Submodule & real & real+syn & real+DR & real+syn+DR \\ \hline 
Cascade R-CNN & HRNetV2p-W32 & - & 25.7 & 25.2 & 25.8 & \textbf{27.1} \\
FoveaBox & ResNeXt101 & FPN+align-gn-ms & 23.7 & - & - & \textbf{26.4}  \\  \hline \hline 		
\multicolumn{3}{c|}{Robotic gastrectomy for gastric cancer} & \multicolumn{4}{c}{mAP $[0.5: 0.05: 0.95]$} \\ \hline 
Model 		& Backbone& Submodule & real & real+syn & real+DR & real+syn+DR \\ \hline 
Cascade R-CNN & HRNetV2p-W32 & - & 38.8 & \textbf{39.8} & 39.7 & 39.6 \\ \
FoveaBox & ResNeXt101 & FPN+align-gn-ms & 37.5 & - & - & \textbf{40.1} \\ \hline
\end{tabular}}
\end{table}

\begin{figure*}[t!]
\includegraphics[width=\textwidth]{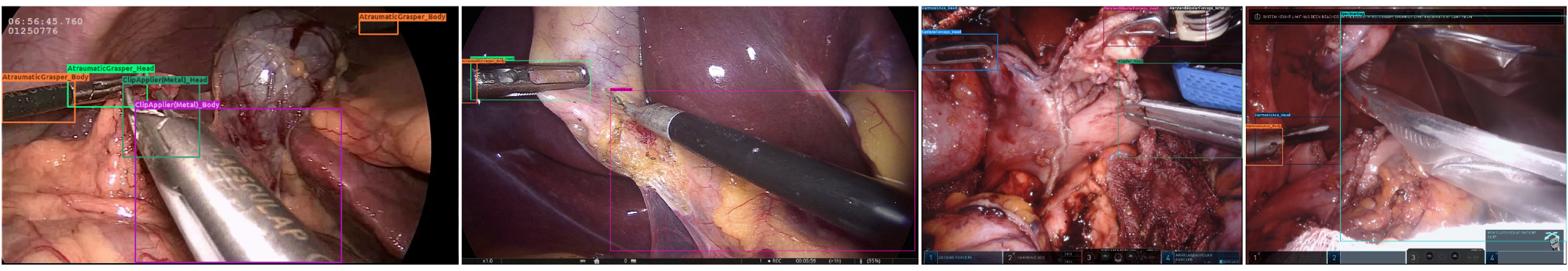}
\caption{\textbf{Visualization of inference results for laparoscopic and robotic surgical tools.} The left two images are the results of Cascade R-CNN for the laparoscopic tools, and the right two images are the results of FoveaBox for the robotic tools.} \label{fig5}
\end{figure*}


\section{Conclusion}
We have published the hSDB-instrument dataset with localization information for the analysis and evaluation of surgical procedures in laparoscopic and robotic surgery. The hSDB-instrument dataset provides bounding box annotations for laparoscopic and robotic surgery, while providing 3D synthetic data to handle class imbalance problems. Besides, the hSDB-instrument dataset is part-specific annotations of the instruments in order to enable the networks to recognize the desired level of subparts of the tools. We also provides the baseline performances of object detection networks for the dataset. We analyzed the hSDB-instrument dataset to provide guidelines for the localization of surgical instruments in surgical videos, while also making the dataset open to the public to contribute to relevant research. We expect that the hSDB-instrument dataset will be of great help in the development of new algorithms related to applications of surgical instrument recognition. \\

\noindent \textbf{Acknowledgement.} This work was supported by the Korea Medical Device Development Fund grant funded by the Korea government (the Ministry of Science and ICT, the Ministry of Trade, Industry and Energy, the Ministry of Health \& Welfare, the Ministry of Food and Drug Safety) (Project Number: 202012A02-02)

%
%

\begin{thebibliography}{99}
\bibitem{Hughes-Hallett15}
Hughes-Hallett, A., Mayer, E. K., Pratt, P. J., Vale, J. A., Darzi, A. W.: Quantitative analysis of technological innovation in minimally invasive surgery. British Journal of Surgery, \textbf{102}(2), 151--157 (2015)
\bibitem{Perez19}
Perez, R. E., Schwaitzberg, S. D.: Robotic surgery: finding value in 2019 and beyond. Annals of Laparoscopic and Endoscopic Surgery,  \textbf{4}(51) (2019)
\bibitem{Jin18}
Jin, A., Yeung, S., Jopling, J., Krause, J., Azagury, D., Milstein, A., Fei-Fei, L.: Tool Detection and Operative Skill Assessment in Surgical Videos Using Region-Based Convolutional Neural Networks. In: Proc. of WACV (2018)
\bibitem{Maier-Hein21}
Maier-Hein, L., Wagner, M., Ross, T. et al.: Heidelberg colorectal data set for surgical data science in the sensor operating room. Sci Data \textbf{8}, 101 (2021).
\bibitem{Twinanda17}
Twinanda, A. P., Shehata, S., Mutter, D., Marescaux, J., de Mathelin, M., Padoy, N.: EndoNet\: A Deep Architecture for Recognition Tasks on Laparoscopic Videos. IEEE Transactions on Medical Imaging, \textbf{36}(1), 86--97 (2017)
\bibitem{Allan19}
Allan, M., Shvets, A., Kurmann, T., Zhang, Z., Duggal, R., Su, Y. H., Rieke, N., Laina, I., Kalavakonda, N., Bodenstedt, S., Garcia-Peraza-Herrera, L. C., Li, W., Iglovikov, V., Luo, H., Yang, J., Stoyanov, D., Maier-Hein, L., Speidel, S., Azizian, M.: 2017 Robotic Instrument Segmentation Challenge. arXiv: 1902.06426 (2019)
\bibitem{Sarikaya17}
D. Sarikaya, J. J. Corso and K. A. Guru.: Detection and Localization of Robotic Tools in Robot-Assisted Surgery Videos Using Deep Neural Networks for Region Proposal and Detection. IEEE Transactions on Medical Imaging, \textbf{36}(7), 1542-1549, (2017),
\bibitem{Ahmidi15}
Ahmidi, N., Tao, L., Sefati, S., Gao, Y., Lea, C., Haro, B. B., Zappella, L., Khudanpur, S., Vidal, R., Hager, G. D.: A Dataset and Benchmarks for Segmentation and Recognition of Gestures in Robotic Surgery. Transaction of Biomedical Engineering, \textbf{64}(9), 2025--2041 (2017)
\bibitem{Lin14}
Lin, T. -Y., Maire, M., Belongie, S., Hays, J., Perona, P., Ramanan, D., Dollar, P., Zitnick, C. L.: Microsoft COCO\: Common Objects in Context. In: Proc. of ECCV (2014)
\bibitem{Geiger12}
Geiger, A., Lenz, P., Urtasun, R.: Are we ready for Autonomous Driving? The KITTI Vision Benchmark Suite In: Proc. of CVPR (2012)
\bibitem{Ren15}
Ren, S., He, K., Girshick, R., Sun, J.: Faster R-CNN\: Towards Real-Time Object Detection with Region Proposal Networks. In: Proc. of NIPS (2015)
\bibitem{Liu18}
Liu, W., Anguelov, D., Erhan, D., Szegedy, C., Reed, S., Fu, C. -Y., Berg, A. C.: SSD\: Single Shot MultiBox Detector. In: Proc. of ECCV (2018)
\bibitem{Lin17-1}
Lin, T. -Y., Dollár, P., Girshick, R., He, K., Hariharan, B., Belongie, S.: Feature Pyramid Networks for Object Detection. In: Proc. of CVPR (2017)
\bibitem{Huang18}
Huang, X., Liu, M. -Y., Belongie, S., Kautz J.: Multimodal Unsupervised Image-to-Image Translation. In: Proc. of ECCV (2018)
\bibitem{Lee19}
Lee, K., Choi, M. -K., Jung, H.: DavinciGAN: Unpaired Surgical Instrument Translation for Data Augmentation. In: Proc. of MIDL (2019)
\bibitem{Park19}
Park, T., Liu, M. -Y., Wang, T. -C., Zhu, J. -Y.: Semantic Image Synthesis with Spatially-Adaptive Normalization. In: Proc. of CVPR (2019)
\bibitem{Pfeiffer19}
Pfeiffer, M., Funke, I., Robu, M. R., Bodenstedt, S., Strenger, L., Engelhardt, Rob, T., Clarkson, M. J., Gurusamy, K., Davidson, B. R., Maier-Hein, L., Riediger, C., Welsch, T., Weitz, J., Speidel S.: Generating large labeled data sets for laparoscopic image processing tasks using unpaired image-to-image translation. In: Proc. of MICCAI (2019)
\bibitem{Chen19}
Chen, K., Wang, J., Pang, J., Cao, Y., Xiong, Y., Li, X., Sun, S., Feng, W., Liu, Z., Xu, J., Zhang, Z., Cheng, D., Zhu, C., Cheng, T., Zhao, Q., Li, B., Lu, X., Zhu, R., Wu, Y., Dai, J., Wang, J., Shi, J., Ouyang, W., Loy, C. C., Lin, D.: MMDetection: Open MMLab Detection Toolbox and Benchmark. arXiv:1906.07155 (2019)
\bibitem{Paszke19}
Paszke, A., Gross, S., Massa, F., Lerer, A., Bradbury, J., Chanan, G., Killeen, T., Lin, Z., Gimelshein, N., Antiga, L., Desmaison, A., Kopf, A., Yang, E., DeVito, Z., Raison, M., Tejani, A., Chilamkurthy, S., Steiner, B., Fang, L., Bai, J., Chintala, S. : PyTorch: An Imperative Style, High-Performance Deep Learning Library. In: Proc. of NeurIPS, 2019.
\bibitem{cvat}
Computer Vision Annotation Tool (CVAT), \url{https://github.com/opencv/cvat}.
\bibitem{Tremblay18}
Tremblay, J., Prakash, A., Acuna, D., Brophy, M., Jampani, V., Anil, C., To, T., Cameracci, E., Boochoon, S., Birchfield, S.: Training Deep Networks with Synthetic Data: Bridging the Reality Gap by Domain Randomization. In: Proc. of CVPRW (2018)
\bibitem{Cai18}
Cai Z., Vasconcelos, N.: Cascade R-CNN: Delving into High Quality Object Detection. In: Proc. of CVPR (2018)
\bibitem{Lin17-2}
Lin, T. -Y., Goyal, P., Girshick, R., He, K., Dollar, P.: Focal Loss for Dense Object Detection. In: Proc. of ICCV (2017)
\bibitem{Tian19}
Tian, Z., Shen, C., Chen, H., He, T.: FCOS: Fully Convolutional One-Stage Object Detection. In: Proc. of ICCV (2019)
\bibitem{Kong19}
Kong, T., Sun, F., Liu, H., Jiang, Y., Shi J.: FoveaBox: Beyond Anchor-based Object Detector. arXiv:1904.03797 (2019)
\bibitem{Pang19}
Pang, J., Chen, K., Shi, J., Feng, H., Ouyang, W., Lin, D.: Libra R-CNN: Towards Balanced Learning for Object Detection. In: Proc. of CVPR (2019)
\bibitem{Li19}
Li, A., Yang, X., Zhang, C.: Rethinking Classification and Localization for Object Detection. In: Proc. of BMVC (2019)
\bibitem{He16}
He, K., Zhang, X., Ren, S., Sun, J.: Deep Residual Learning for Image Recognition. In: Proc. of CVPR (2016)
\bibitem{Xie17}
Xie, S., Girshick, R., Dollár, P., Tu, Z., He K.: Aggregated Residual Transformations for Deep Neural Networks. In: Proc. of CVPR (2017)
\bibitem{Wang19}
Wang, J., Sun, K., Cheng, T., Jiang, B., Deng, C., Zhao, Y., Liu, D., Mu, Y., Tan, M., Wang, X., Liu, W., Xiao, B.: Deep High-Resolution Representation Learning for Visual Recognition. arXiv:1908.07919 (2019)
\end{thebibliography}
%

\end{document}